\documentclass[journal]{IEEEtran}
\usepackage{times}
\usepackage{epsfig}
\usepackage{graphicx}
\usepackage{amsmath}
\usepackage{amssymb}
\usepackage{helvet}
\usepackage{courier}
\usepackage{stfloats}
\ifCLASSINFOpdf
\else
\fi
\begin{document}
\title{A Novel Method to Study Bottom-up Visual Saliency and Its Neural Mechanism}
%
%
%

\author{Cheng~Chen,
        Xilin~Zhang,
        Yizhou~Wang,
        and~Fang~Fang}

%
%

\markboth{The manuscript has been uploaded on ArXiv}
{Shell \MakeLowercase{\textit{et al.}}: Bare Demo of IEEEtran.cls for Journals}
%



\maketitle

\begin{abstract}
In this study, we propose a novel method to measure bottom-up saliency maps of natural images. In order to eliminate the influence of top-down signals, backward masking is used to make stimuli (natural images) subjectively invisible to subjects, however, the bottom-up saliency can still orient the subjects¡¯ attention. To measure this orientation/attention effect, we adopt the cueing effect paradigm by deploying discrimination tasks at each location of an image, and measure the discrimination performance variation across the image as the attentional effect of the bottom-up saliency. Such attentional effects are combined to construct a final bottomup saliency map. Based on the proposed method, we introduce a new bottom-up saliency map dataset of natural images to benchmark computational models. We compare several state-of-the-art saliency models on the dataset. Moreover, the proposed paradigm is applied to investigate the neural basis of the bottom-up visual saliency map by analyzing psychophysical and fMRI experimental results. Our findings suggest that the bottom-up saliency maps of natural images are constructed in V1. It provides a strong scientific evidence to resolve the long standing dispute in neuroscience about where the bottom-up saliency map is constructed in human brain.
\end{abstract}


\section{Introduction}
Visual attention is essential for many cognitive processes, e.g. scene recognition. Information selection in these processes can be executed by top-down signals voluntarily, or triggered by salient stimuli in a bottom-up way. The former is driven by specific goals, while the latter is guided by a bottom-up saliency map. A bottom-up saliency map is defined as a topographical map to describe and predict the distribution of attention attraction on a visual input \cite{Koch}. It is able to efficiently direct our focus of attention. Hence, understanding bottom-up saliency not only has a scientific value, but also can help us to build intelligent yet computation efficient models to facilitate real world applications such as object detection in clustered environment and unmanned vehicles (\cite{Rutishause,Serre}).

Many bottom-up saliency models have been proposed to simulate human intelligence in visual attention. Compared to the numerous proposed bottom-up visual saliency computational models, methods about how to accurately measure/quantize bottom-up saliency on natural images are under-studied. Currently, in order to evaluate computational saliency models, researchers use eye tracking data to generate groundtruth, e.g. \cite{Bruce,Wang2}. In traditional methods, participants are asked to free view an image presented on a monitor for a short period of time. After that, fixations of each participant are extracted by certain computational algorithms \cite{Bruce} from the original eye tracking data. And then, by fusing all the fixations across participants, researchers build a saliency benchmark dataset.

However, using eye tracking data to build a bottom-up saliency benchmark dataset has several disadvantages. First, eye fixations are mostly used to study overt attention. They are very easily to be influenced by top-down signals and center bias, and they are considered to be lack of precision and have large individual variances. Second, eye fixations are estimated by computational algorithms from raw eye tracking dataset, hence, the saliency values derived from the estimated fixations could be easily influenced by different algorithms. Therefore, it is not ideal/appropriate to use eye fixation data as the groundtruth to evaluate bottom-up saliency models. To solve this problem, it is imperative to propose a rigorous method to measure the bottom-up saliency map free from top-down signals.

In this paper, we propose a new method to measure bottom-up saliency maps of natural images. In the proposed paradigm, top-down control is blocked using {\em backward masking} method. In this way, stimuli (natural images) are made subjectively invisible to subjects, however, bottom-up saliency can still orient the subjects' attention. To measure this orientation/attention effect, we adopt the {\em cueing effect paradigm} by deploying discrimination tasks at each location of an image, and measure the discrimination performance variation across the image as the attentional effect of the bottom-up saliency. Such attentional effects are integrated to construct a final bottom-up saliency map. Based on the proposed method, we introduce a new bottom-up saliency map dataset of natural images to benchmark computational models. Moreover, we evaluate several state-of-the-art saliency models using the dataset.

To justify the proposed paradigm, it is applied to investigate the neural basis of the bottom-up visual saliency by analyzing psychophysical and fMRI experimental results, especially focusing on identifying the brain areas that construct the bottom-up saliency maps of natural images. Currently, there are controversies about where the bottom-up saliency maps are constructed in human brain. Many brain regions that are believed to realize bottom-up saliency maps receive both bottom-up and top-down signals, and the judgement could be made based on the mixed signals rather than the pure bottom-up one. By using the proposed paradigm, top-down signals can be effectively eliminated, which makes it possible to identify the brain areas that realize the bottom-up saliency maps.

We design novel psychophysical and physiological experiments to locate the brain areas that construct bottom-up visual saliency. In the psychophysical experiment, by using the proposed method, we measure the attentional effect induced by bottom-up saliency. In the physiological experiment, we record the blood-oxygen-level-dependent (BOLD) signals responding to the subjectively invisible natural images using functional magnetic resonance imaging (fMRI). We find that when top-down signals are eliminated, (i) the attentional effect increases with the degree of saliency, (ii) the BOLD signals increase with the degree of saliency only in V1, but not in other cortical areas such as V4, IPS or FEF, and more importantly, (iii) the attentional effect of bottom-up saliency significantly correlates with the BOLD signal in V1 across subjects. These findings suggest that {\em bottom-up saliency is constructed in V1.}, which challenges the argument that bottom-up saliency can only be constructed in higher cortical areas such as IPS or FEF (\cite{Bogler,Geng}).

In summary, the proposed has the following contributions. (i) We develop a novel method to measure bottom-up saliency maps of natural images, and construct a dataset to benchmark bottom-up saliency models more precisely. (ii) The research result is a step toward resolving a fundamental scientific dispute about where the bottom-up saliency is constructed in brain. (iii) The finding also suggests a clear reference to implement biologically-plausible computational models of bottom-up visual saliency, for example, by simulating the function of V1 neurons.

\subsection{Related Work}
There is a huge body of research focusing on studying competing computational bottom-up saliency models. However, few of them investigate a proper method to measure the bottom-up saliency maps of natural images in computer vision field. Bruce and his colleagues suggest to evaluate the computed saliency maps by using eye fixations collected in an image viewing task \cite{Bruce}. In their method, subjects are asked to observe an image presented on a CRT monitor without any particular instructions. For each image, a fixation density map is produced by convolving a 2D Gaussian function with each fixation collected in this image. The generated fixation density map could be further used to compare with a computed saliency map, so that the performance of a computational saliency model can be evaluated.
Many other studies adopt the eye fixation dataset proposed by Bruce \cite{Bruce} to evaluate their own models, e.g., \cite{Duan,Hou,Wang}. However, using eye fixation data to infer the distribution of bottom-up attention is far from  satisfaction. There are two major drawbacks. First, eye fixations are very sparse. The saliency values of the locations not visited by eye fixations are not measurable; they can only be estimated via smoothing or set to 0 as in \cite{Bruce}, which can be inappropriate. Second, as aforementioned, eye fixations can be easily influenced by top-signals \cite{Corbetta} or the prior knowledge/experience of individuals. Therefore, the attentional effect resulted from the bottom-up saliency is hard to differentiate. Therefore, to evaluate/benchmark bottom-up saliency models, it is imperative to propose a new method to measure saliency maps of natural images derived from pure bottom-up signals and it should be able to measure saliency values at each location of an image.

To block the top-down signals, several methods can be used to achieve this goal, such as continuous flash suppression \cite{Fang} and backward masking \cite{Breitmeyer}. In the paradigm of backward masking, a stimulus is presented very briefly and followed by a high contrast mask, so that subjects cannot be aware of the stimulus \cite{Zhang}. The attentional attraction can be measured by using a modified version of the cueing effect paradigm proposed by Posner et al. \cite{Posner}. In this paradigm, a target appears in one of two locations in the peripheral visual field randomly, and subjects need to perform a discrimination task with this target. Before the target presentation, a cue will indicate the location of the following target. Trials with a correct cue belong to the valid cue condition, while trials with an incorrect cue belong to the invalid cue condition. The attentional attraction could be measured by comparing the performance (response time or accuracy) between the two conditions \cite{Eckstein}.

Such a paradigm can also be used to solve the controversy about the neural basis of bottom-up saliency maps. Several studies had found that different brain regions could realize the saliency map. Subcortical structures such as superior colliculus \cite{Fecteau} and pulvinar \cite{Shipp} were found to be able to construct the saliency map. Mazer and Gallant found that extrastriate ventral area V4 could realize a retinotopic saliency map to guide eye movement, providing evidence that brain activities in the ventral pathway could be correlated with the saliency map \cite{Mazer}. Their conclusion was also supported by Asplund and his colleagues, who found that the ventral network can account for stimulus-driven attention \cite{Asplund}. Geng and Mangun found that the anterior intraparietal sulcus (aIPS) was sensitive to the bottom-up influence driven by stimulus saliency \cite{Geng}. Besides, frontal eye fields (FEF) was also found to play an important role in decoding the winner-take-all (WTA) stage of saliency processing \cite{Bogler}. Most of these findings were consistent with a dominant view, which argues that the final saliency map results from pooling different visual features channels after each visual feature channel construct its salient map independently. Thus, higher brain regions could be the possible candidates that realize the saliency map. However, these higher brain regions receive both bottom-up and top-down signals \cite{Gilbert2}, it is hard to justify that the pure bottom-up saliency is constructed there.

On the other hand, Li proposed the V1 theory, which argues that neural activities in V1 could create the bottom-up saliency map via intracortical interactions that are manifested in contextual influences (\cite{Li1}, \cite{Li2}). Several studies also support the V1 theory. By measuring reaction times searching for a singleton that differs from its surrounding in more than one feature, Koene and Zhaoping found their results were consistent with the properties of some V1 neurons \cite{Koene}, which provides evidence for the V1 hypothesis of the bottom-up saliency map. Furthermore, Zhang and his colleagues measured the attentional effect of the bottom-up saliency map of simple texture stimuli, and found the degree of attention attraction correlated with the amplitude of the earliest component of the ERP as well as the V1 BOLD signal across subjects \cite{Zhang}, which also supported the V1 hypothesis.

An important reason that the controversy exists is that most of these brain regions receive both bottom-up and top-down signals, which makes it difficult to determine whether the saliency map they realized truly reflected the bottom-up attention attraction or not. In order to investigate the neural basis of the bottom-up saliency map, it is important to probe bottom-up signals free from top-down signals. Based on this requirement, Zhang and his colleagues used simple oriented bars as their stimuli, and found their results supported the V1 hypothesis \cite{Zhang}. However, as we know, neurons in V1 are highly tuned to these oriented bars.  Such difference in stimulus preference between V1 and other cortical areas makes their conclusion not so convincing \cite{Betz}. Moreover, our world contains much more complex visual features, thus the V1 hypothesis should be verified on natural scenes.

The rest of this paper is organized as follows. In Section 2, we introduce the material and method to measure the bottom-up saliency maps. The comparison results among different existing saliency models are presented in Section 3. In Section 4, we introduce our psychophysical and fMRI experiments to investigate the neural basis of the bottom-up saliency map of natural images. The results of these experiment are also presented in Section 5. Finally, we conclude and discuss the paper in Section 6.

\section{Measuring the bottom-up saliency map of natural images}
In this section, we will introduce the protocol of our method to measure the bottom-up saliency map of natural images, followed by some descriptions about the visual stimuli and participants.

\subsection{Experimental protocol}
We perform psychophysical experiments measure bottom-up saliency maps from human. In the main experiment, visual stimuli are presented on a Gamma-corrected Iiyama HM204DT 22 inches monitor, with a spatial resolution of $1024 \times 768$ and a refresh rate of 60Hz. The viewing distance is 83cm. Participants' head positions are stabilized using a chin rest and a head rest. A white fixation cross is always presented at the center of the monitor, and participants are asked to fixate the cross throughout the experiment.

\begin{figure}[t]
\begin{center}
   \includegraphics[width=1\linewidth]{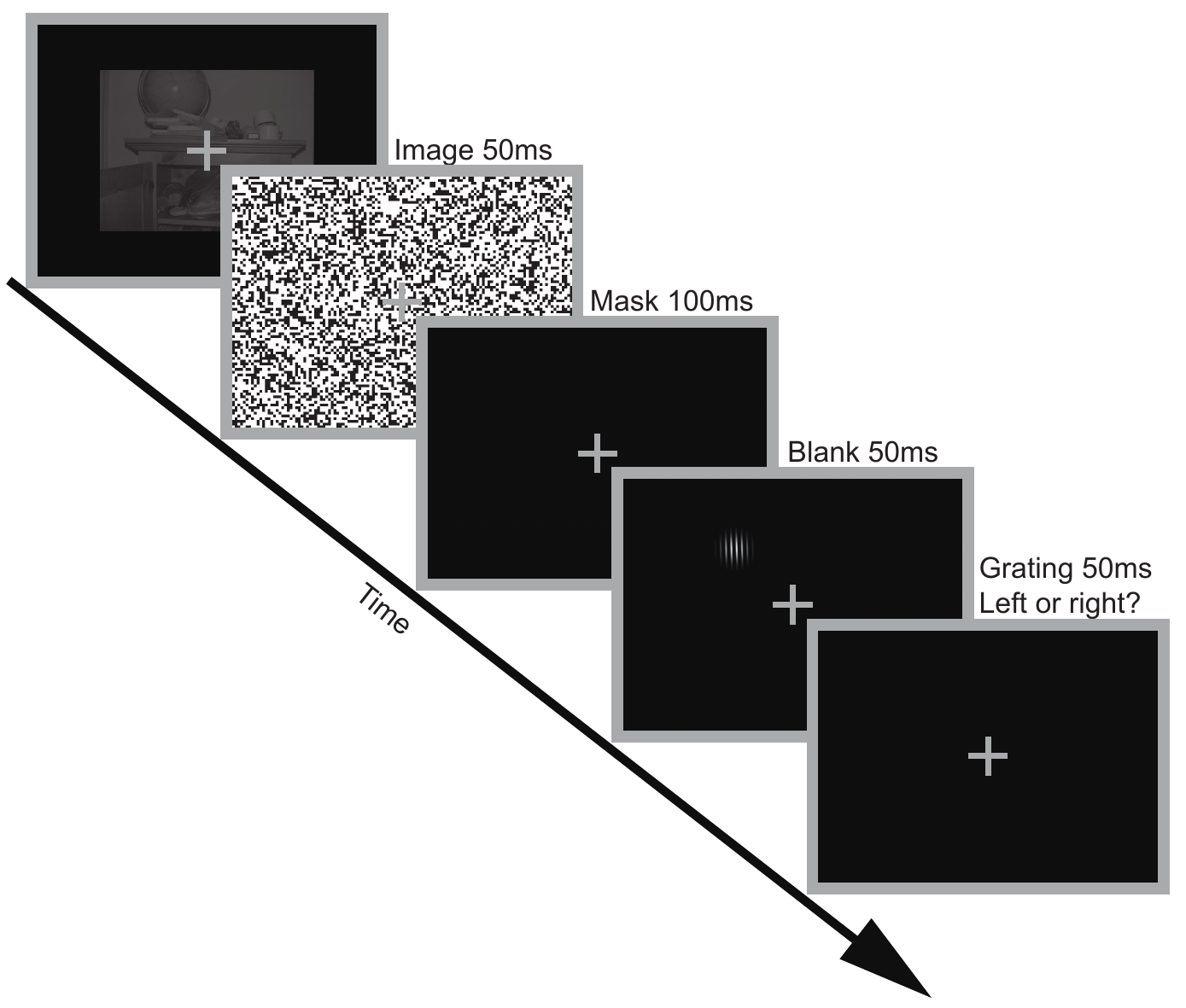}
\end{center}
   \caption{The psychophysical procedure to measure the bottom-up saliency map of natural images.}
\label{fig:1}
\end{figure}

To measure the bottom-up saliency map of a natural image, a low-luminance image (namely the {\em image condition}) or a blank (namely the {\em baseline condition}) is presented at the center of the screen for 50 ms, followed by a 100 ms full screen mask, and then 50 ms fixation interval. The bottom-up saliency map of the image serves as a cue to attract spatial attention, and the mask could ensure the image subjectively invisible to participants \cite{Breitmeyer}. Then a Gabor orientates at $\pm$ 1.5$^\circ$ away from the vertical is presented at a random location within the area of the image. Participants are asked to press one of the two keys to indicate the orientation of the Gabor (left or right tilted). The duration of each trial is 2s. Fig.~\ref{fig:1} shows the protocol of a trial. Each {\em condition} consists of 10 blocks of 96 trials. The location and the orientation of the Gabors are counterbalanced across trials.

At each location of an image, we measure the performance variation of the orientation discrimination task between the image and the baseline conditions. Such difference indicates the degree of attentional attraction due to the bottom-up saliency. Compared with the {\em baseline condition}, locations in the {\em image condition} with higher bottom-up saliency will attract more attention to improve the orientation discrimination task performance, while other locations will attract less attention so that the performance at these locations have less improvement.

Moreover, the mask following the image could ensure that top-down signals are eliminated from the experiment \cite{Breitmeyer}. To ensure this precondition, subjects are asked to complete a two-alternative forced choice (2AFC) experiment in a criterion-free way before the main experiment. The protocol of this experiment is similar to the main experiment. Each trial begins with either a low-luminance image or a blank, followed by a full screen mask. Participants are asked to make a forced choice response to judge whether there is an image presented before the mask or not. We adjust the image luminance so as to make the subjects' performance at chance level, this ensures that the masked images are indeed subjectively invisible, consequently, eliminate the top-down signals in the experiment.

\subsection{Visual stimuli}
There are three kinds of visual stimuli used in our experiment: natural images, masks and Gabors.

\textbf{Natural images}: 40 representative natural images are selected from the Internet and several public datasets. They cover a variety of image categories including ocean, mountain, open-country, street, tower, wild animal, etc. All these images are scaled to the same size ($480 \times 640$ pixels, $14.54^\circ \times 19.38^\circ$ of visual angle) with a mean low luminance (2.9 cd/m$^2$). Such low luminance is aimed to ensure that these images could be easily masked so that top-down signals could be eliminated.

\textbf{Mask}: Mask stimuli are high contrast checkerboards with randomly arranged checkers. The size of each checker is about $0.25^\circ\times0.25^\circ$. The luminance of black checkers is 1.8 cd/m$^2$, and the luminance of white checkers is 79 cd/m$^2$.

\textbf{Gabor}: Gabor is a kind of visual pattern that human visual system is sensitive to. In our experiment, the Gabor has a spatial frequency of 5.5 cpd (cycle per degree) and its diameter is $2.5^\circ$ with full contrast. The Gabor orientates at  $\pm$ 1.5$^\circ$ away from the vertical.

\subsection{Participants}
 Twenty human subjects (10 female and 10 male, 18-26 years old) participate in the experiment. All participants are na\"{\i}ve to the purpose of the study. All of them are right-handed, report normal or corrected-to-normal vision, and have no known neurological or visual disorders. They give written, informed consent in accordance with the procedures and protocols approved by the human subjects review committee of Peking University.

\section{A dataset of the bottom-up saliency map of natural images}

As mentioned above, using eye tracking data as the groundtruth of bottom-up saliency maps is not appropriate. The proposed method provides a viable means \-- it is able to measure the bottom-up saliency value at any position of an image without being affected by top-down signals. Based on the proposed method, we propose a more precise bottom-up saliency map dataset of natural images, so that bottom-up saliency models can be benchmarked.

In our experiment, each image is partitioned into an imaginary $6\times 8$ grid. Although such resolution might be low, compared with traditional eye tracking methods, the density of the sample points is larger and our measurement covers the whole image. Moreover, the resolution can be increased by adding more sample points. We measure the saliency value at the center of each cell and assign the value to the rest pixels within the cell. We normalize the saliency values to [0, 1] on each image. Finally, the obtained map is convolved with a 2D Gaussian function to generate a smooth bottom-up saliency map. For each image, we collect data from 20 subjects. We apply this measurement with 40 natural images to build a bottom-up saliency map dataset.

\begin{figure}[t]
\begin{center}
   \includegraphics[width=1\linewidth]{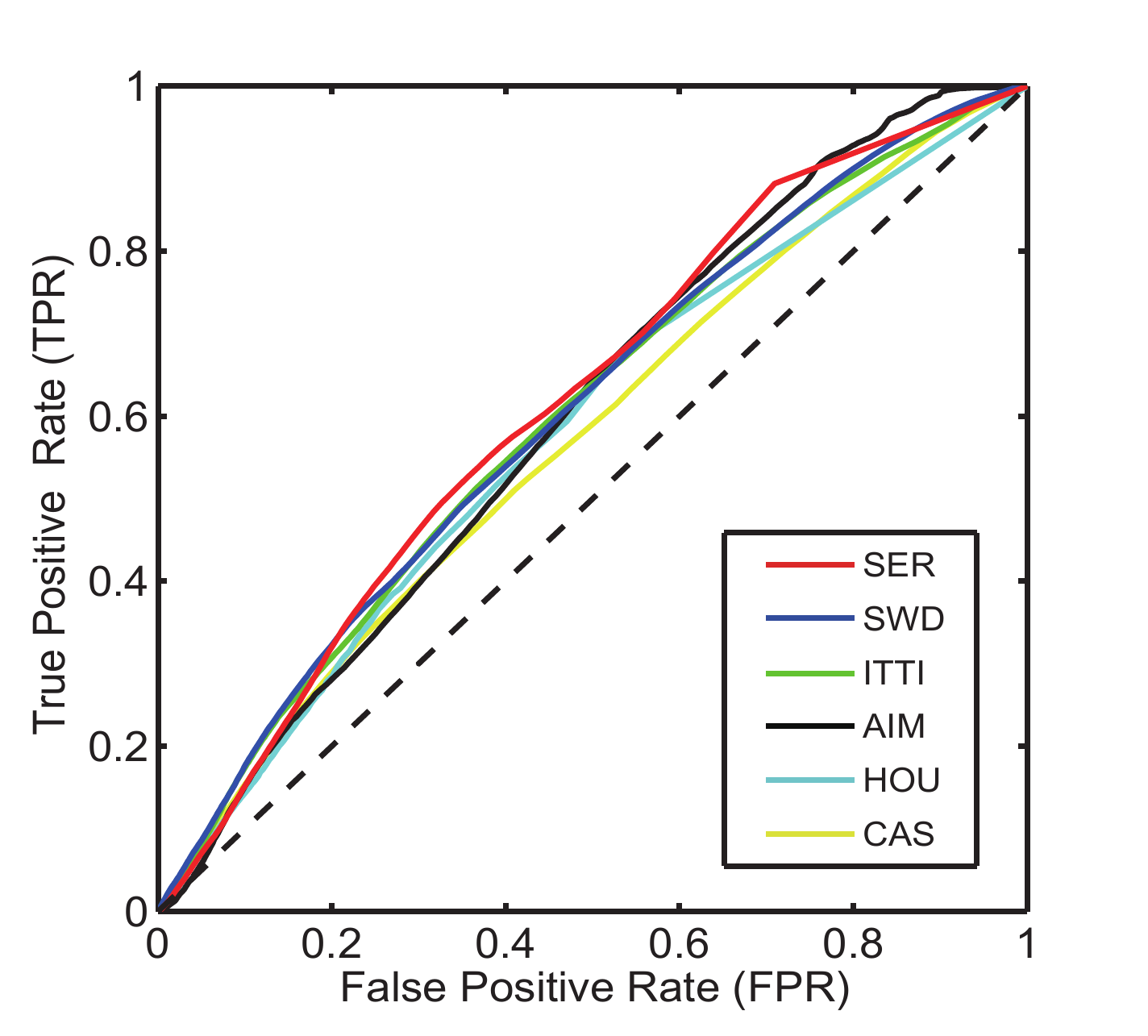}
\end{center}
   \caption{The ROC curves of several saliency models on our dataset.
   }
\label{fig:2}
\end{figure}

We compare the Receiver Operator Characteristic (ROC) curves and the ROC areas of several state-of-the-art bottom-up computational saliency models: AIM \cite{Bruce}, SWD \cite{Duan}, CAS \cite{Goferman}, HOU \cite{Hou}, ITTI \cite{Itti1}, and SER \cite{Wang} on our dataset. The reason we select these saliency models is that they have public code. In the ROC analysis, the values of the measured saliency maps are normalized to [0, 1], and we use 0.5 as the threshold to binarize the saliency maps to salient regions and non-salient ones as the ground truth. To evaluate the computed saliency maps (also normalized), we adjust the threshold from 0 to 1, and identify the {\em false positives} and {\em true positives} at each threshold, then generate the ROC curves. The ROC curves are shown in Fig.~\ref{fig:2} and the ROC areas are shown in Table ~\ref{table1}. The larger the ROC area, the better.

\begin{table}[ht]
\caption{The ROC areas comparison on our dataset}
\label{table1}
\begin{center}
\begin{tabular}{|c|c|}
\hline
Saliency models &ROC area\\
\hline
SER \cite{Wang} &0.6149 \\
\hline
SWD \cite{Duan} &0.6036 \\
\hline
ITTI \cite{Itti1} &0.6003 \\
\hline
AIM \cite{Bruce} &0.5997 \\
\hline
HOU \cite{Hou} &0.5824 \\
\hline
CAS \cite{Goferman} &0.5725 \\
\hline
\end{tabular}
\end{center}
\end{table}

It can be seen that several bottom-up saliency models, such as ITTI \cite{Itti1}, SWD \cite{Duan}, and SER \cite{Wang}) achieves better performance on our dataset. ITTI \cite{Itti1} is a traditional bottom-up saliency model based on the center-surround mechanism. SWD \cite{Duan} is another bottom-up saliency model based on the spatially weighted dissimilarity. SER \cite{Wang} simulates the signals transmission among the inter-connected V1 neurons, which is another bottom-up saliency model. Moreover, it is noticeable that the context-aware saliency model (CAS) proposed by Goferman et al. \cite{Goferman}, which is based on {\em top-down signals}, has a poor performance on this dataset. More importantly, the settings of our experiment is convenient, e.g. our method does not require eye-tracking equipment, so that researchers could easily measure the bottom-up saliency map of natural images and evaluate their saliency models.


\section{Neural basis of the bottom-up saliency map of natural images}
Our paradigm could block the top-down signals, and thus also provide an efficient means to investigate the neural basis of the bottom-up saliency map of natural images. In our study, psychophysical and fMRI experiments using a similar paradigm are performed. In this section we will introduce the participants,  the visual stimuli used in psychophysical and fMRI experiments, as well as the experimental procedures and data analysis processes for these two experiments.

\subsection{Participants}
Sixteen human subjects (9 female and 7 male, 19-26 years old) participate in the experiment. All of them participate in the psychophysical experiment. Thirteen of them participate in the fMRI experiment. Two subjects in the fMRI experiment are excluded from data analysis because of excessive head motion during fMRI scanning. All subjects are na\"{\i}ve to the purpose of the study except for one subject (one of the authors). All of them are right-handed, report normal or corrected-to-normal vision, and have no known neurological or visual disorders. They give written, informed consent in accordance with the procedures and protocols approved by the human subjects review committee of Peking University.

\subsection{Visual stimuli}
\begin{figure}
\begin{center}
\includegraphics[width=1\linewidth]{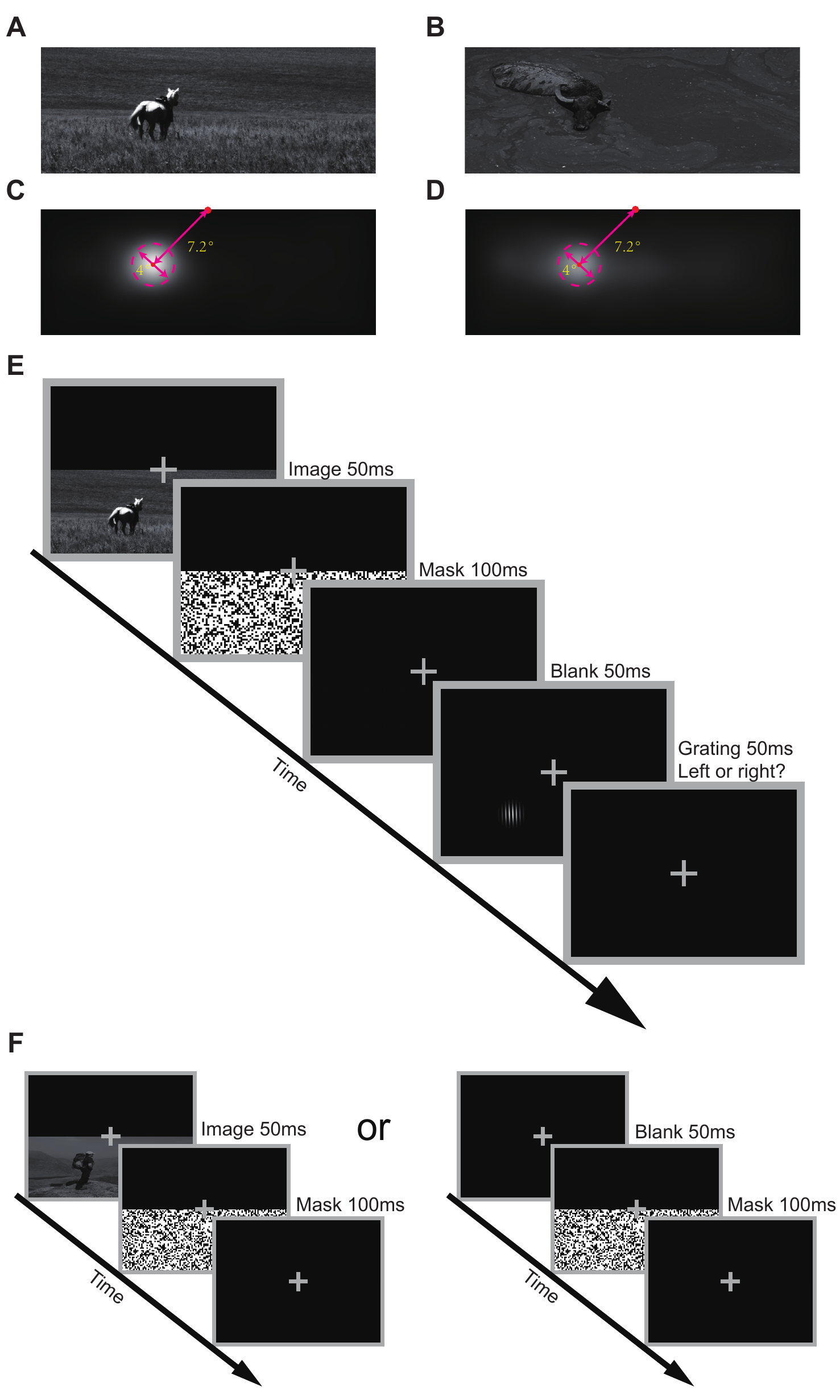}
\end{center}
   \caption{Stimuli and psychophysical experiment procedure. (A and B) Examples of high salient (A) and low salient (B) natural images. (C and D) The averaged saliency maps of 25 high salient images (C) and 25 low salient images (D) with a salient region left to the fixation. Areas with a high luminance level have a high saliency.  (E) Psychophysical protocol to measure the attentional effect of the bottom-up saliency maps of natural images. (F) The procedure of the additional 2AFC experiment to confirm that natural images are invisible to participants in our paradigm.}
\label{fig:3}
\end{figure}

To study the neural basis of bottom-up visual saliency, the visual stimuli used in the psychophysical and fMRI experiments must be carefully processed. First we collect a large number of natural images from the Internet and several public datasets. All these images are scaled to the same size ($11.63^\circ\times31.03^\circ$ of visual angle) with a mean low luminance (2.9 cd/m$^2$) (Fig.~\ref{fig:3}A and Fig.~\ref{fig:3}B). Then we calculate the computational saliency map of each image by using the bottom-up visual saliency model proposed by Itti and his colleagues \cite{Itti1}. Then, we select 50 images for the current experiment. Each of these images has only one round salient region centers at about 7.2$^\circ$ eccentricity in the lower left quadrant (left salient images). The diameter of the salient region is about 4$^\circ$ (Fig.~\ref{fig:3}C and Fig.~\ref{fig:3}D). Moreover, to measure the bottom-up saliency map quantitatively, we classify all images into two groups: the high salient group and the low salient group, based on the proposed saliency index in the following formulation:

\begin{equation}
  Index(n) = \frac{S_{I}(n)-S_{O}(n)}{S_{O}(n)},
\end{equation}

where $n$ denotes the index of an image. $S_{I}$ denotes the averaged saliency value of the round salient region in Fig.~\ref{fig:3}C and Fig.~\ref{fig:3}D, and $S_{O}$ denoted the averaged saliency value of the residual region (the rest of the image). The higher value of the $Index$, the higher saliency of the round region relative to the rest of the image. We selected half of images with an upper 50\% $Index$ as the high salient image set, and the rest as the low salient image set. An example of the high salient and the low salient images can be found in Fig.~\ref{fig:3}A and Fig.~\ref{fig:3}B. The averaged saliency maps of the 25 high salient images is shown in Fig.~\ref{fig:3}C, and the low salient counterpart is shown Fig.~\ref{fig:3}D.

In order to balance the location of the salient region, each of the left salient images is flipped horizontally to generate a new image, which has a salient region in the lower right quadrant (right salient images). Notice that the contents of the left and the right salient images are the same, the only difference between them is the location of the salient region. Thus, the stimuli used in the psychophysical and the fMRI experiments consist of two groups: the high salient group and the low salient group. Each group contains 50 images, half of which are the left salient images and the other half are the right salient ones. In both experiments, these stimuli are presented in the lower visual field on a dark screen (1.8 cd/m$^2$).

Mask stimuli are high contrast checkerboards with randomly arranged checkers (Fig.~\ref{fig:3}E). The size of each checker is about $0.25^\circ\times0.25^\circ$. The luminance of black checkers is 1.8 cd/m$^2$, while the luminance of white checkers is 79 cd/m$^2$.

\subsection{Experimental procedure}

\textbf{Psychophysical experiment}:Each trial starts with a fixation at the center of the screen. A natural image is presented on the lower half of the screen for 50 ms, followed by a 100 ms mask at the same position, and another 50 ms fixation interval.  Then a Gabor centers at about $7.2^\circ$ eccentricity from the fixation is presented randomly at either the lower left quadrant or the lower right quadrant with equal probability for 50 ms. The location of the Gabor is either at the salient region of the preceding image or its contralateral counterpart, thus indicating the {\em valid cue condition} or the {\em invalid cue condition}. The Gabor has a spatial frequency of 5.5 cpd (cycle per degree) and its diameter is $2.5^\circ$ with full contrast. The Gabor orientates at  $\pm$ 1.5$^\circ$ away from the vertical. Subjects are asked to press one of the two keys to indicate the orientation of the Gabor (left or right tilted). The duration of each trial is 2s.

Fig.~\ref{fig:3}E shows the procedure of our experiment. The experiment consists of 10 runs. Each run contains 100 trials with two conditions: the high salient condition and the low salient condition. The images for the two conditions are randomly selected from the high and the low salient image groups correspondingly. The attentional effect of bottom-up saliency maps for each condition is quantified as the difference between the orientation discrimination task performance (discrimination accuracy) of the {\em valid cue condition} and the {\em invalid cue conditions}.

Moreover, we also ask subjects to complete an additional two-alternative forced choice (2AFC) experiment to check whether these natural images are subjectively invisible to subjects in our paradigm. In the additional experiment, each trial begin with either a low-luminance image or a blank, followed by a mask. Subjects are asked to make a forced choice response to judge whether there is an image presented before the mask or not (Fig.~\ref{fig:3}F). The performance at chance level in this experiment could provide an objective confirmation that the masked images are indeed invisible.

\textbf{fMRI experiment}: The fMRI experiment is conducted to investigate the neural basis of the bottom-up saliency map. In the fMRI experiment, an event-related design is adopted. The experiment consists of eight functional scans of 125 continuous trials. Each scan begins with 6s fixation and lasts 268s. There are five types of trials, including four types of salient image trials: two degrees of saliency (high and low) $\times$ two locations of salient region (left and right), and fixation trials. In all types of trials, a white fixation cross is always presented at the center of the screen. In a salient image trial, an image is presented on the lower half of the screen for 50 ms, followed by a 100 ms mask at the same position and 1850 ms fixation. Subjects are asked to indicate the location of the salient region, which is left to the fixation in half of salient image trials and right in the other half randomly. In a fixation trial, only the fixation point is presented for 2000ms. In a scan, there are 25 trials for each type. The order of the trials is counterbalanced across eight scans using M-sequences \cite{Buracas}. These are pseudo random sequences that have the advantage of being perfectly counterbalanced n trial back, so that each type of trials is preceded and followed equally often by all types of trials, including itself.

Retinotopic visual areas (V1, V2, V3 and V4) are defined by a standard phase-encoded method developed by Sereno et al. \cite{Sereno} and Engel et al. \cite{Engel}, in which subjects view rotating wedge and expanding ring stimuli that create traveling waves of neural activity in visual cortex. A block-design scan is used to define the regions of interest (ROIs) in LGN, V1-V4, LOC, IPS and FEF corresponding to the salient region. The scan consists of twelve 12s stimulus blocks, interleaved with twelve 12s blank intervals. In a stimulus block, subjects passively view images with colorful natural scenes, which have the same size as the salient region in the natural images, and are presented at the location of the salient region (either left or right to the fixation). Images appear at a rate of 8Hz.

\begin{figure}
\begin{center}
\includegraphics[width=0.8\linewidth]{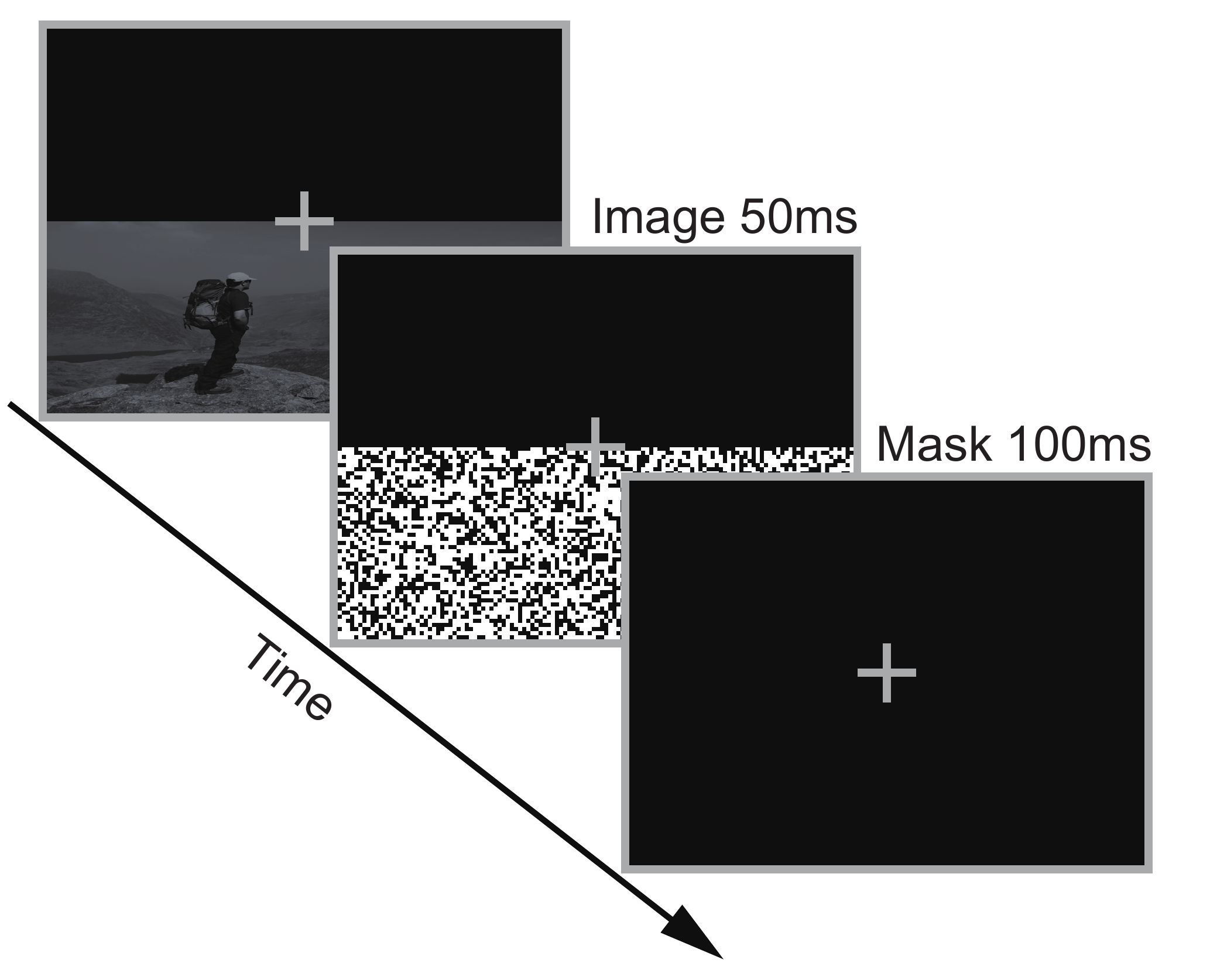}
\end{center}
   \caption{The procedure of the fMRI experiment.}
   \label{fig:4}
\end{figure}

\subsection{MRI data acquisition}

In the scanner, the stimuli are back-projected via a video projector (refresh rate: 60Hz; spatial resolution: $1024 \times 768$) onto a translucent screen placed inside the scanner bore. Subjects viewe the stimuli through a mirror located above their eyes. The viewing distance is 83cm. MRI data are collected using a 3T Siemens Trio scanner with a 12-channel phase-array coil. Blood oxygen level-dependent (BOLD) signals are measured with an echo-planar imaging sequence (TE: 30ms; TR: 2000ms; FOV: $186 \times 192 mm^2$; matrix: $62 \times 64$; flip angle: 90; slice thickness: 5mm; gap: 0mm; number of slices: 30; slice orientation: coronal). The fMRI slices cover the occipital lobe, most of the parietal lobe and part of the temporal lobe. A high-resolution 3D structural data set (3D MPRAGE; $1 \times 1 \times 1 mm^3$ resolution) is collected in the same session before the functional runs. All the subjects undergo with two sessions, one for the retinotopic mapping and the other for the main experiment.

\subsection{MRI data processing and analysis}

The anatomical volume for each subject in the retinotopic mapping session is transformed into the AC-PC (anterior commissure-posterior commissure) space and then inflated using Brain Voyager QX. Functional volumes in all the sessions for each subject are preprocessed, including 3D motion correction, linear trend removal, and high-pass (0.015 Hz) filtering using Brain Voyager QX. Head motion within any fMRI session is \textless 2mm for all subjects except two subjects excluded for further analysis because of excessive head motion. fMRI images are then aligned to the anatomical volume in the retinotopic mapping session and transformed into the AC-PC space. The first 6s of BOLD signals are discarded to minimize transient magnetic saturation effects.

A general linear model (GLM) procedure is used for ROI analysis. The ROIs in LGN, V1-V4, LOC, IPS and FEF are defined by a localizer scan and retinotopic mapping scans (p \textless 10$^{-8}$, uncorrected). Talairach coordinates of rLGN, lLGN, rFEF and lFEF are ($21\pm3$, $-26 \pm 3$, $-1 \pm 3$), ($-23 \pm 3$, $-27 \pm 2$, $-1 \pm 2$), ($37 \pm 7$, $-10 \pm 5$, $47 \pm 4$) and ($-37 \pm 6$, $-12 \pm 4$, $46 \pm 3$) respectively, consistent with previous studies (e.g. \cite{Chen}, \cite{Paus}).

The event-related BOLD signals are calculated separately for each subject, following the method used by Kourtzi and Kanwisher \cite{Kourtzi}. For each event-related scan, the time course of the MR signal intensity is first extracted by averaging the data from all the voxels within the predefined ROI. The average event-related time course is then calculated for each type of trial, by selectively averaging from stimulus onset and using the average signal intensity during the fixation trials as a baseline to calculate percent signal change. Specifically, in each scan we average the signal intensity across the trials for each type of trial at each of 12 corresponding time points starting from the stimulus onset. These event-related time courses of the signal intensities are then converted to time courses of percent signal change for each type of trials by subtracting the corresponding value for the fixation trials and then dividing by that value. Because M-sequences have the advantage that each type of trials was precede and followed equally often by all types of trials, the overlapping BOLD responses due to the short interstimulus interval are removed by this averaging procedure \cite{Buracas}. The resulting time course for each type of trials is then averaged across scans and subjects.

\subsection{Computational Saliency Model}
We adopt an influential computational saliency model proposed by Itti et al. \cite{Itti1} to measure the bottom-up saliency map of each image. The model is based on the center-surround mechanism, and combined information from three channels: color, intensity and orientation. By using this model, we could predicate the degree of saliency of each image based on the formulation we proposed.

\section{The Results of the neural basis of the bottom-up saliency maps}

\subsection{Psychophysical experiment}
16 subjects participate in this experiment. In the additional 2AFC experiment, subjects report that they are unaware of the natural images. The percentages of correct detection (mean $\pm$ SEM) are $48.6 \pm 1.5\%$ and $50.9 \pm 1.4\%$ for the high salient and the low salient groups, respectively. The results are statistically indistinguishable from chance level, which indicates that the natural images in both two groups are subjectively invisible to subjects.

\begin{figure}[t]
\begin{center}
   \includegraphics[width=0.8\linewidth]{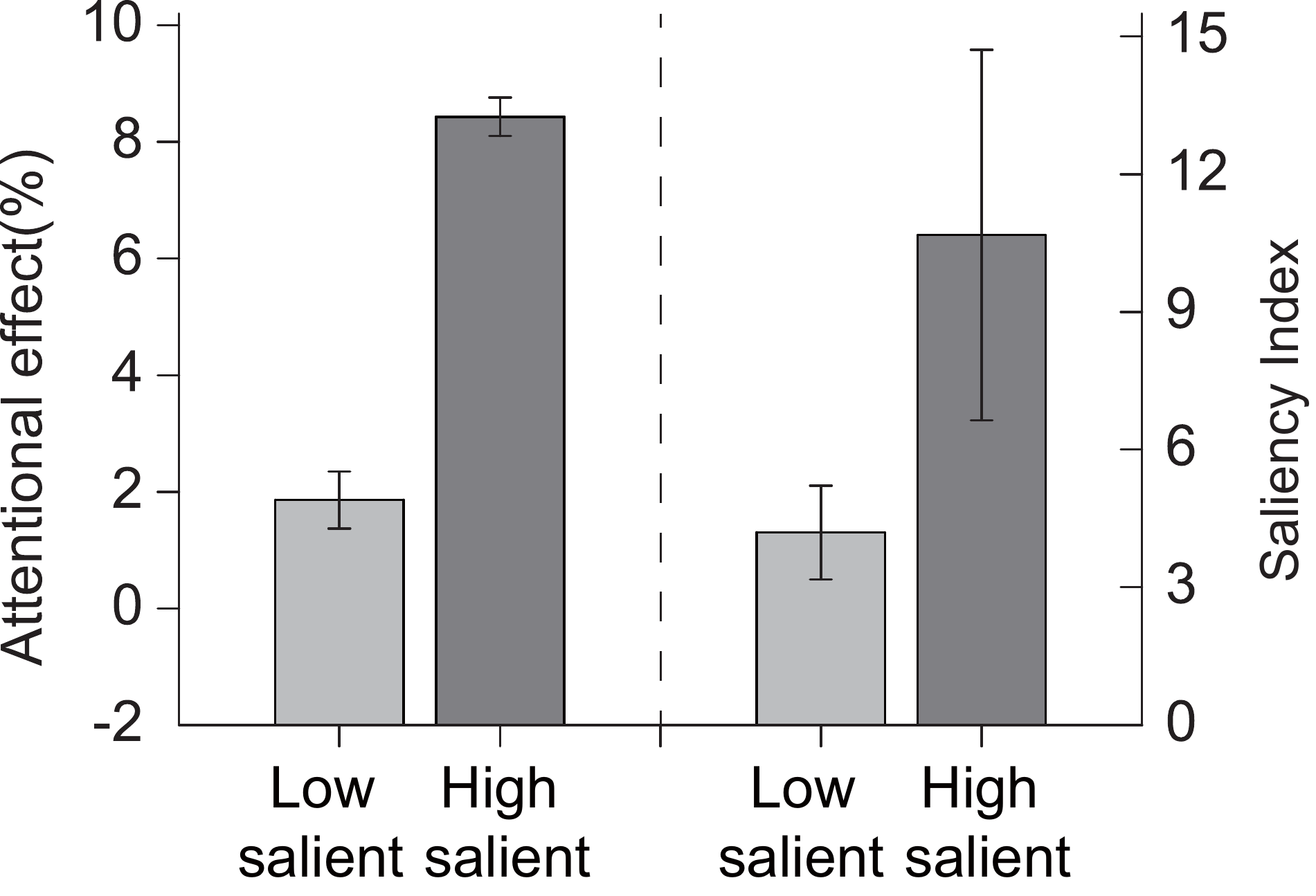}
\end{center}
   \caption{Attentional effects in psychophysical data and from computational saliency model prediction. The left two bars indicate the psychophysical attentional effects for the high salient and the low salient group, respectively. Error bars denote 1 SEM across subjects in each group. The right two bars indicate the saliency index calculated by a computational saliency model \cite{Itti1} for the high salient and the low salient group, respectively. Error bars denote 1 SD across all images in each group. The saliency index predicts the degree of attentional attraction.}
\label{fig:5}
\end{figure}

\begin{figure*}
\begin{center}
   \includegraphics[width=0.8\linewidth]{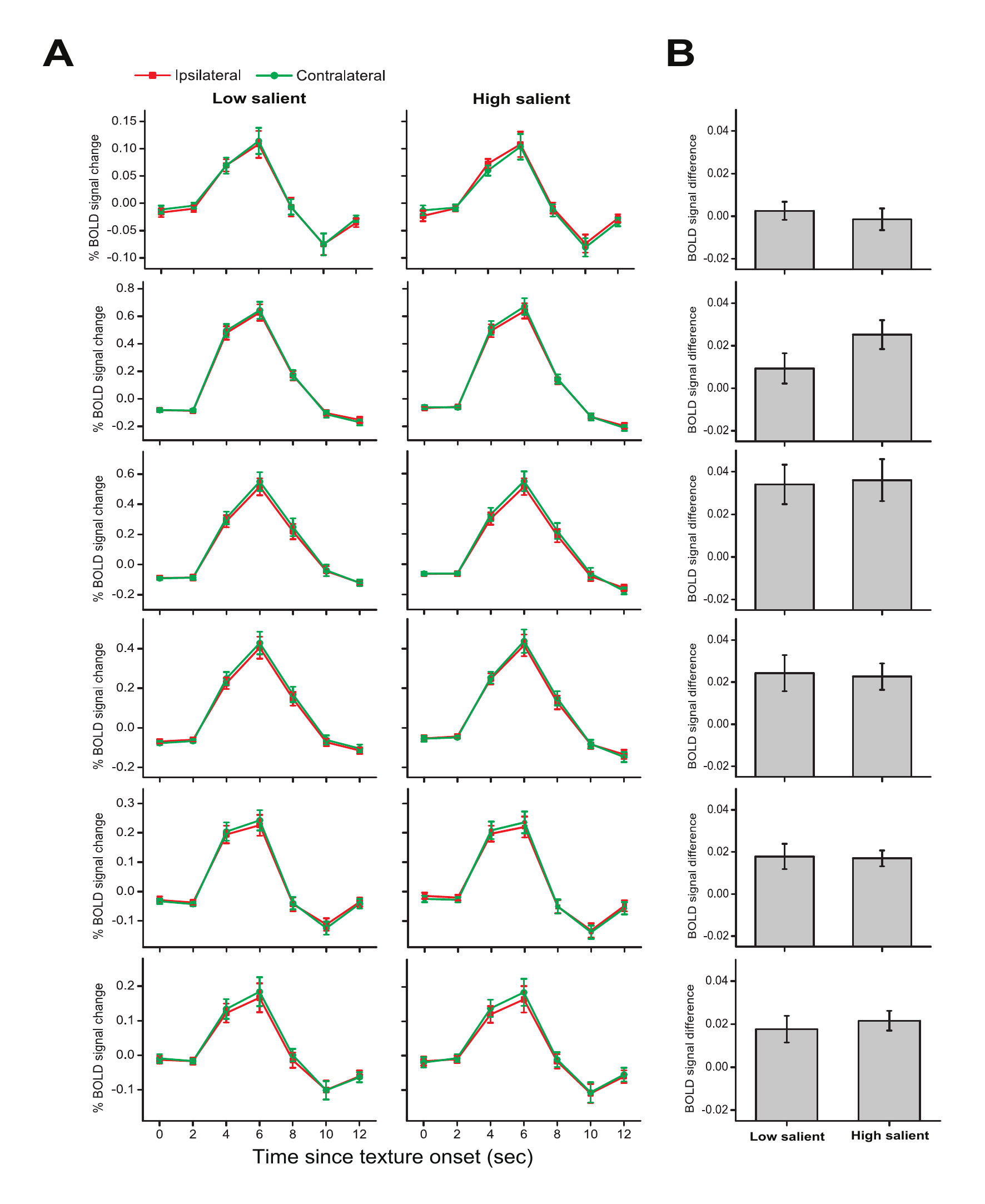}
\end{center}
   \caption{FMRI results in LGN, V1-V4 and IPS.
   (A). Event-related BOLD signals averaged across subjects in the contralateral and ipsilateral ROIS in LGN, V1-V4 and IPS. They are evoked by the bottom-up saliency map of natural images in the high salient and the low salient group. Error bars denote 1 SEM calculated across subjects at each time point.
(B) Peak amplitude differences between the event-related BOLD signal in the contralateral ROI and that in the ipsilateral ROI in LGN, V1-V4 and IPS for the high salient and the low salient group. Error bars denote 1 SEM calculated across subjects.
}
\label{fig:6}
\end{figure*}

In the main experiment, considering that the salient region of a natural image could serve as a cue to attract attention, the attentional effect of the bottom-up saliency map of invisible natural images is quantified as the difference between the accuracy of the Gabor orientation discrimination performance in the {\em valid cue condition}, and that in the {\em invalid cue condition}. We find that the discrimination accuracy is higher in the {\em valid cue condition} than that in the {\em invalid cue condition}, for both high salient images (Valid: $81.31\pm0.98\%$; Invalid $72.88 \pm 0.98\%$) and low salient images (Valid: $77.86 \pm 0.93\%$; Invalid $76.54 \pm 0.88\%$). Thus, the attentional effect of the bottom-up saliency maps for the high salient and low salient groups are $8.43 \pm 0.33\%$ and $1.32 \pm 0.47\%$, respectively (left panel in Fig.~\ref{fig:6}). The results indicate that the bottom-up saliency maps exhibit a positive cueing effect even when the image is subjectively invisible, which suggests that subjects¡¯ attention is attracted to the salient region of the invisible images, so that they perform better in the valid cue condition than in the invalid cue condition. Moreover, the attentional effect in both two groups is significant (paired t-test high salient: $t_{15} = 18.126,\ p < 0.001$; low salient: $t_{15} = 2.782,\ p = 0.014$; significant level $\alpha  = 0.05$). The attentional effect of the high salient images is also significantly greater than that of the low salient images ($t_{15} = 9.665,\ p < 0.001$). We also calculated the saliency index of the high salient and the low salient images based on the proposed formulation. The saliency index predicted the degree of the attention attraction of a bottom-up saliency map (right plane in Fig.~\ref{fig:6}). Psychophysical data were consistent with the prediction from the computational model.

\subsection{fMRI Experiment}

\begin{figure}
\begin{center}
   \includegraphics[width=1\linewidth]{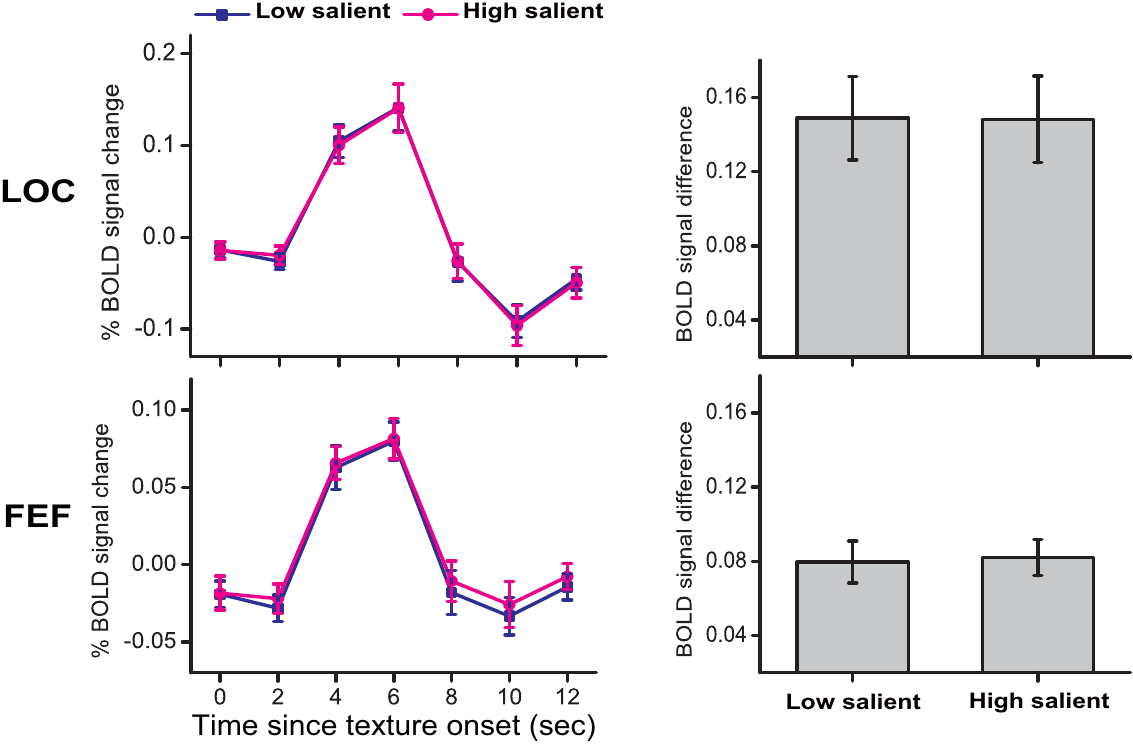}
\end{center}
   \caption{FMRI results in LOC and FEF.
Left column: event-related BOLD signals averaged across subjects in LOC and FEF. They were evoked by the bottom-up saliency map of natural images in the high salient and the low salient group. Error bars denote 1 SEM calculated across subjects at each time point.
Right column: Peak amplitude of the event-related BOLD signals in LOC and FEF for the high salient and the low salient group. Error bars denote 1 SEM calculated across subjects.
}
\label{fig:7}
\end{figure}

\begin{figure*}
\begin{center}
   \includegraphics[width=0.7\linewidth]{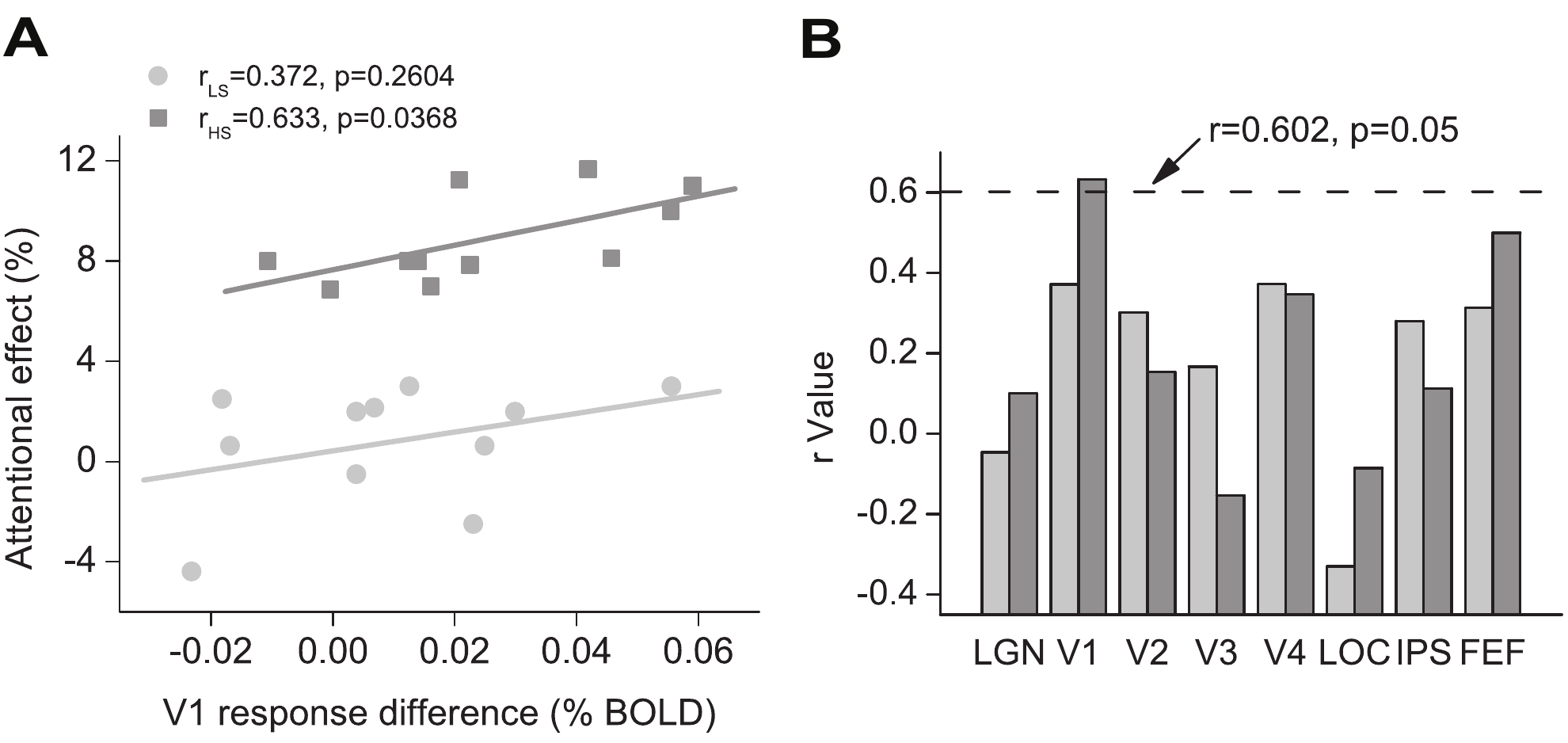}
\end{center}
   \caption{Correlations between the psychophysical and the fMRI measures across individual subjects.
    (A) Correlations between the attentional effect and the BOLD signal difference in V1 for the high salient (dark gray) and the low salient (light gray) group.
(B) Correlation coefficients (the $r$ values) between the attentional effect and the BOLD signal difference in LGN, V1-V4, LOC, IPS and FEF for the high salient (dark gray) and the low salient (light gray) group.
}
\label{fig:8}
\end{figure*}

11 subjects participate in this experiment. The percentages of correct detection are $50.1 \pm 0.7\%$ and $50.3 \pm 0.7\%$ for the high salient and the low salient groups, respectively. The behavioral data confirms that the low-luminance natural images are indeed subjectively invisible to subjects. Contralateral and ipsilateral regions of interest (ROIs) in LGN, V1-V4 and IPS are defined as the cortical areas that respond to retinal inputs in the salient region and its contralateral counterpart. LOC and FEF in two hemispheres could be activated equally well by stimuli presented in the left and the right visual fields in the localizer scan. Thus, instead of presenting data in ipsilateral and contralateral regions in LOC and FEF, we directly analyze event-related BOLD signals according to the degree of saliency (the high salient and the low salient groups) in these two cortical areas.

It is found that in V1-V4 and IPS, the natural images in both two groups evoke larger BOLD signals in the contralateral ROIs compared with the ipsilateral ROIs (Fig.~\ref{fig:4}A). It means that the salient region could evoke stronger neural activity than its contralateral counterpart. BOLD signal difference (namely {\bf BSD}) is quantified as the peak value difference of the BOLD signals in the contralateral ROI and that in the ipsilateral ROI (Fig.~\ref{fig:4}B). The {\bf BSD}s of the high salient group and the low salient group are compared by {\em paired t-test}. We find that the {\bf BSD} of the high salient group is significantly higher than that of the low salient group ($t_{10}=4.989,\ p=0.001$) in V1. However, in LGN, V2-V4 and IPS, we do NOT observe significant {\bf BSD} difference between the high salient group and the low salient group (LGN: $t_{10}=-0.690,\ p=0.506$; V2: $t_{10}=0.194,\ p=0.850$; V3: $t_{10}=0.194,\ p=0.850$; V4: $t_{10}=-0.159,\ p=0.877$; IPS: $t_{10}=0.540,\ p=0.601$). Moreover, we also measure the BOLD signal peak value difference between the high salient and the low salient groups in LOC and FEF (Fig.~\ref{fig:7}). We find NO significant difference in these two areas (LOC: $t_{10}=-0.141,\ p=0.891$; FEF: $t_{10}=-0.690,\ p=0.506$). As the attentional effect of the bottom-up saliency map of the high salient group is also significantly higher than that of the low salient group, these findings reveal that neural activity in V1 is parallel to the attentional effect.

\subsection{Correlation Analysis}

In order to further evaluate the role of neural activities of early visual cortical areas in realizing bottom-up saliency maps, we calculate the correlation coefficients between our psychophysical and fMRI measures across individual subjects. The attentional effect is significantly correlated with the BOLD signal difference inV1 for the high salient group ($r=0.633,\ p<0.05$), but not for the low salient group ($r=0.372,\ p=0.260$) (Fig.~\ref{fig:8}A). However, no significant correlation is found between the attentional effect and the BOLD signal difference in the other cortical areas (Fig.~\ref{fig:8}B). The results indicate a close relationship between the attentional effect and V1 neural activities.

\section{Conclusion and Discussion}

In summary, we develop a novel method to measure the bottom-up saliency map of natural images. Our paradigm is able to measure the bottom-up saliency at every location of an image without affected by top-down signals, which remedies several drawbacks of the traditional eye tracking based methods. Based on the proposed method, a new database of the bottom-up saliency maps of natural images is built, which provides a more precise measurement to benchmark computational saliency models. Several state-of-art saliency models are compared on the proposed dataset. Moreover, a similar paradigm is applied to investigate the neural basis of the bottom-up saliency map of natural images. We measure the attentional effect and brain activities of the bottom-up saliency map on subjectively invisible natural images. We find that even if subjects are unaware of natural images, the attentional effect and the BOLD signal in V1 still increase with the degree of saliency. In addition, the attentional effect significantly correlates with the BOLD signals in V1, but not other cortical areas. These findings provide a strong scientific evidence to resolve the long standing dispute in neuroscience about where the bottom-up saliency map is constructed in human brain.

One important assumption of our study is that top-down signals are effectively eliminated in our paradigm. In the area of cognitive neuroscience, it has been proved and widely accepted that subjective awareness is determined by top-down signaling ~\cite{Del}. Thus, rendering a stimulus invisible means eliminating top-down signals to the stimulus. No matter in psychophysical or fMRI experiments, we find that those natural images are invisible to participants, which confirms the assumption that top-signals are indeed eliminated in our paradigm. This issue is quiet important because several studies indicated that temporarily sluggish fMRI signals reflect neural activities resulting from both bottom-up and top-down processes, even in early visual cortex \cite{Harrison}. Thus, blocking top-down signals make sure that we could observe a relative pure signal of the bottom-up saliency map of natural scenes in different brain regions.

Another important concern about our method is the resolution of the measured bottom-up saliency map. It can be laborious to measure the saliency value at each location/pixel of an image. However, we can propose a coarse-to-fine strategy to increase the sampling resolution non-uniformly. For example, we can re-sample more points around the salient regions based on the previous pass of sampling.

The most interesting observation in our experiment is that we find V1 plays a distinct role in realizing the bottom-up saliency map of natural scenes. Our findings challenge the dominant view that the bottom-up saliency map is created in higher brain regions such as IPS and FEF (\cite{Koch}, \cite{Geng}, \cite{Bogler}). Neural activities induced by the bottom-up saliency map are not observed in LOC, IPS, or FEF, which indicates that the observed neural activities in V1 are not attribute to the signals feedback from these areas. On the other hand, our results are consistent with previous findings that the neural response of V1 neurons is higher when their preferred stimuli pop out from the background \cite{Marcus}. We suggest that the underlying neural mechanism of our observation may be attributed to the lateral connections \cite{Gilbert} between V1 neurons. More importantly, our observation is consistent with the ¡®V1 saliency theory¡¯ (\cite{Li1}, \cite{Li2}), which states that V1 creates the bottom-up saliency map. Our experiments extend previous evidences which support this theory from both psychophysical and physiological aspects. Instead of using textures consisted of simple oriented bars, we use natural scenes as our stimuli. Natural scenes contain richer naturalistic low-level features, including luminance, contrast, spatial frequency, curve etc. These features are basic units that our visual system needs to deal with. Therefore, this study is not only a critical complement to the previous study, but also provides an important evidence for the V1 saliency map argument.

%
\IEEEpeerreviewmaketitle


%

%

\ifCLASSOPTIONcaptionsoff
  \newpage
\fi



%



\bibliographystyle{ieee}
\bibliography{egbib}

\begin{thebibliography}{10}\itemsep=-1pt

\bibitem{Asplund}
C.~Asplund, J.~Todd, A.~Snyder, and R.~Marois.
\newblock A central role for the lateral prefrontal cortex in goal-directed and
  stimulus-driven attention.
\newblock {\em Nature Neuroscience}, 13(4):507--514, 2010.

\bibitem{Betz}
T.~Bezt, N.~Wilming, C.~Bogler, and J.~Haynes.
\newblock Dissociation between saliency signals and activity in early visual
  cortex.
\newblock {\em Journal of Vision}, 13(14):1--12, 2013.

\bibitem{Bogler}
C.~Bogler, S.~Bode, and J.~Haynes.
\newblock Decoding successive computational stages of saliency processing.
\newblock {\em Current Biology}, 21(19):1667--1671, 2011.

\bibitem{Breitmeyer}
B.~G. Breitmeyer and H.~Ogmen.
\newblock Recent models and findings in visual backward masking: A comparison,
  review, and update.
\newblock {\em Perception and Psychophysics}, 62(8):1572--1595, 2000.

\bibitem{Bruce}
N.~Bruce and J.~Tsotsos.
\newblock Saliency based on information maximization.
\newblock {\em Advances in Neural Information Processing Systems}, 18:155--162,
  2006.

\bibitem{Buracas}
G.~Buracas and G.~Boynton.
\newblock Efficient design of event-related fmri experiments using m-sequences.
\newblock {\em Neuroimage}, 16:801--813, 2002.

\bibitem{Chen}
W.~Chen, X.~Zhu, K.~Thulborn, and K.~Ugurbil.
\newblock Retinotopic mapping of lateral geniculate nucleus in humans using
  functional magnetic resonance imaging.
\newblock {\em Proceedings of the National Academy of Sciences of the United
  States of America}, 96(5):2430--2434, 1999.

\bibitem{Corbetta}
M.~Corbetta and G.~L. Shulman.
\newblock Control of goal-directed and stimulus-driven attention in the brain.
\newblock {\em Nature Review Neuroscience}, 3(3):201--215, 2002.

\bibitem{Del}
A.~D. Cul, S.~Baillet, and S.~Dehaene.
\newblock Brain dynamics underlying the nonlinear threshold for access to
  consciousness.
\newblock {\em PLoS Biology}, 5(10):2408--2423, 2007.

\bibitem{Duan}
L.~Duan, C.~Wu, J.~Miao, L.~Qing, and Y.~Fu.
\newblock Visual saliency detection by spatially weighted dissimilarity.
\newblock {\em IEEE Computer Vision and Pattern Recognition}, pages 473--480,
  2011.

\bibitem{Eckstein}
M.~Eckstein, S.~Schimozaki, and C.~Abbey.
\newblock The footprints of visual attention in the posner cueing paradigm
  revealed by classification images.
\newblock {\em Journal of Vision}, 2(1):25--45, 2002.

\bibitem{Engel}
S.~Engel, G.~Glover, and B.~Wandell.
\newblock Retinotopic organization in human visual cortex and the spatial
  precision of functional mri.
\newblock {\em Cerebral Cortex}, 7(2):181--192, 1997.

\bibitem{Fang}
F.~Fang and S.~He.
\newblock Cortical responses to invisible objects in the human dorsal and
  ventral pathways.
\newblock {\em Nature Neuroscience}, 8(10):1380--1385, 2005.

\bibitem{Fecteau}
J.~Fecteau and D.~Munoz.
\newblock Salience, relevance, and firing: a priority map for target selection.
\newblock {\em Trends in Cognitive Science}, 10(9):25--45, 2006.

\bibitem{Geng}
J.~Geng and G.~Mangun.
\newblock Anterior intraparietal sulcus is sensitive to bottom-up attention
  driven by stimulus salience.
\newblock {\em Journal of Cognitive Neuroscience}, 21(8):1584--1601, 2009.

\bibitem{Gilbert2}
C.~Gilbert and W.~Li.
\newblock Top-down influences on visual processing.
\newblock {\em Nature Reviews Neuroscience}, 14:350--363, 2013.

\bibitem{Gilbert}
C.~Gilbert and T.~Wiesel.
\newblock Clustered intrinsic connections in cat visual cortex.
\newblock {\em Journal of Neuroscience}, 3(5):1116--1133, 1983.

\bibitem{Goferman}
S.~Goferman, L.~Zelnik-Manor, and A.~Tal.
\newblock Context-aware saliency detection.
\newblock {\em IEEE Transaction in Pattern Analysis and Machine Intelligence},
  34(10):1915--1926, 2012.

\bibitem{Harrison}
S.~Harrison and F.~Tong.
\newblock Biologically plausible saliency mechanisms improve feedforward object
  recognition.
\newblock {\em Vision Research}, 50(22):2295--2307, 2010.

\bibitem{Hou}
X.~Hou and L.~Zhang.
\newblock Dynamic visual attention: Searching for coding length increments.
\newblock {\em Advances in Neural Information Processing Systems}, 21:681--688,
  2008.

\bibitem{Itti1}
L.~Itti, C.~Koch, and E.~Niebur.
\newblock A model of saliency based visual attention for rapid scene analysis.
\newblock {\em IEEE Transaction in Pattern Analysis and Machine Intelligence},
  20(11):1254--1259, 1998.

\bibitem{Koch}
C.~Koch and S.~Ullman.
\newblock Shifts in selective visual attention: towards the underlying neural
  circuitry.
\newblock {\em Human Neurobiology}, 4(4):219--227, 1985.

\bibitem{Koene}
A.~Koene and L.~Zhaoping.
\newblock Feature-specific interactions in salience from combined feature
  contrasts: Evidence for a bottom-up saliency map in v1.
\newblock {\em Journal of Vision}, 7(7):1--14, 2007.

\bibitem{Kourtzi}
Z.~Kourtzi and N.~Kanwisher.
\newblock Cortical regions involved in perceiving object shape.
\newblock {\em Journal of Neuroscience}, 20(9):3310--3318, 2000.

\bibitem{Li1}
Z.~Li.
\newblock Contextual influences in v1 as a basis for pop out and asymmetry in
  visual search.
\newblock {\em Proceedings of the National Academy of Sciences of the United
  States of America}, 96(18):10530--10535, 1999.

\bibitem{Li2}
Z.~Li.
\newblock A saliency map in primary visual cortex.
\newblock {\em Trends in Cognitive Science}, 6(1):9--16, 2002.

\bibitem{Marcus}
D.~Marcus and D.~V. Essen.
\newblock Scene segmentation and attention in primate cortical areas v1 and v2.
\newblock {\em Journal of Neurophysiology}, 88(5):2648--2658, 2002.

\bibitem{Mazer}
J.~Mazer and J.~Gallant.
\newblock Goal-related activity in v4 during free viewing visual search:
  Evidence for a ventral stream visual salience map.
\newblock {\em Neuron}, 40(6):1241--1250, 2003.

\bibitem{Paus}
T.~Paus.
\newblock Location and function of the human frontal eye field: a selective
  review.
\newblock {\em Neuropsychologia}, 34(6):475--483, 1996.

\bibitem{Posner}
M.~Posner, C.~Snyder, and B.~Davidson.
\newblock Attention and the detection of signals.
\newblock {\em Journal of Experimental Psychology}, 109(2):160--174, 1980.

\bibitem{Rutishause}
U.~Rutishause, D.~Walther, C.~Koch, and P.Perona.
\newblock Is bottom-up attention useful for object recognition?
\newblock {\em IEEE Computer Vision and Pattern Recognition}, 2:37--44, 2004.

\bibitem{Sereno}
M.~Sereno, A.~Dale, J.~Reppas, K.~Kwong, J.~Belliveau, T.~Brady, B.~Rosen, and
  R.~Tootell.
\newblock Borders of multiple visual areas in human revealed by
  functionalmagnetic resonance imaging.
\newblock {\em Science}, 268(5212):889--893, 1995.

\bibitem{Serre}
T.~Serre, L.~Wolf, and T.~Poggio.
\newblock Object recognition with features inspired by visual cortex.
\newblock {\em IEEE Computer Vision and Pattern Recognition}, 2:994--1000,
  2005.

\bibitem{Shipp}
S.~Shipp.
\newblock The brain circuitry of attention.
\newblock {\em Trends in Cognitive Sciece}, 8(5):223--230, 2004.

\bibitem{Wang2}
W.~Wang, C.~Chen, Y.~Wang, T.~Jiang, F.~Fang, and Y.~Yao.
\newblock Simulating human saccadic scanpaths on natural images.
\newblock {\em IEEE Computer Vision and Pattern Recognition}, pages 441--448,
  2011.

\bibitem{Wang}
W.~Wang, Y.~Wang, Q.~Huang, and W.~Gao.
\newblock Measuring visual saliency by site entropy rate.
\newblock {\em IEEE Computer Vision and Pattern Recognition}, pages 2368--2375,
  2010.

\bibitem{Zhang}
X.~Zhang, L.~Zhaoping, T.~Zhou, and F.~Fang.
\newblock Neural activities in v1 create a bottom-up saliency map.
\newblock {\em Neuron}, 73(1):183--192, 2012.

\end{thebibliography}

%








\end{document}